%% file: main.tex
\documentclass[10pt,twocolumn,letterpaper]{article}

\usepackage[pagenumbers]{cvpr} 

\usepackage[utf8]{inputenc} 
\usepackage[T1]{fontenc}    
\usepackage{url}            
\usepackage{booktabs}       
\usepackage{amsfonts}       
\usepackage{nicefrac}       
\usepackage{microtype}      
\usepackage{xcolor}         
\usepackage{comment}

\usepackage{comment}
\usepackage{graphicx}
\usepackage{adjustbox}
\usepackage[most]{tcolorbox} 
\usepackage{multirow}
\usepackage{arydshln}
\usepackage[dvipsnames]{xcolor}
\usepackage{array}
\usepackage{float}

\usepackage[noend]{algpseudocode}
\usepackage{algorithm}

\usepackage{amsmath}
\usepackage{amssymb}
\usepackage{booktabs}

\usepackage{epsfig}
\usepackage{boldline}
\usepackage{booktabs}

\usepackage{amsthm}

\theoremstyle{definition}

\lstdefinestyle{mystyle}{
    language=Python,
    basicstyle=\footnotesize\ttfamily, 
    breaklines=true,                
    keywordstyle=\color{blue},        
    commentstyle=\color{green!40!black},
    stringstyle=\color{purple},       
    numbers=left,                   
    numberstyle=\tiny\color{gray},  
    showstringspaces=false,         
    tabsize=2,                      
    frame=single
}
\input{preamble}
\definecolor{cvprblue}{rgb}{0.21,0.49,0.74}
\usepackage[pagebackref,breaklinks,colorlinks,allcolors=cvprblue]{hyperref}

\def\paperID{****} 
\def\confName{****}
\def\confYear{****}

\title{ObjectAlign: Neuro-Symbolic Object Consistency Verification and Correction}

\author {
    Mustafa Munir\textsuperscript{\rm 1},
    Harsh Goel\textsuperscript{\rm 1},
    Xiwen Wei\textsuperscript{\rm 1},
    Minkyu Choi\textsuperscript{\rm 1}, 
    Sahil Shah\textsuperscript{\rm 1}, \\
    Kartikeya Bhardwaj\textsuperscript{\rm 2},
    Paul Whatmough\textsuperscript{\rm 2},
    Sandeep Chinchali\textsuperscript{\rm 1},
    Radu Marculescu\textsuperscript{\rm 1}
    \\
    The University of Texas at Austin\textsuperscript{\rm 1}, Qualcomm AI Research\textsuperscript{\rm 2}\\
    \small
    \texttt{\{mmunir, harshg99, xiwenwei, minkyu.choi, ss96869, sandeepc, radum\}@utexas.edu},  \\
    \small
    \texttt{\{kbhardwa, pwhatmou\}@qti.qualcomm.com}
}

\normalsize

\begin{document}
\maketitle

\newcommand{\truedensity}{p_\mathrm{data}}
\newcommand{\truedistr}{P_\mathrm{data}}
\newcommand{\denoiser}{\epsilon_\theta}

\newcommand{\andlogic}{\wedge}
\newcommand{\orlogic}{\vee}
\newcommand{\notlogic}{\neg}
\newcommand{\imply}{\implies}

\newcommand{\always}{\Box}
\newcommand{\eventually}{\diamondsuit}
\newcommand{\ournext}{\mathsf{X}}
\newcommand{\until}{\mathsf{U}}
\begin{abstract}
Video editing and synthesis often introduce object inconsistencies, such as frame flicker and identity drift that degrade perceptual quality. To address these issues, we introduce ObjectAlign, a novel framework that seamlessly blends perceptual metrics with symbolic reasoning to detect, verify, and correct object-level and temporal inconsistencies in edited video sequences. The novel contributions of ObjectAlign are as follows: First, we propose learnable thresholds for metrics characterizing \emph{object consistency} (i.e. CLIP-based semantic similarity, LPIPS perceptual distance, histogram correlation, and SAM-derived object–mask IoU). Second, we introduce a neuro-symbolic verifier that combines two components: (a) a formal, SMT-based check that operates on masked object embeddings to provably guarantee that object identity does not drift, and (b) a temporal fidelity check that uses a probabilistic model checker to verify the video's formal representation against a temporal logic specification ($\Phi$). A frame transition is subsequently deemed “consistent” based on a single logical assertion that requires satisfying both the learned metric thresholds and this unified neuro-symbolic constraint, ensuring both low-level stability and high-level temporal correctness. Finally, for each contiguous block of flagged frames, we propose a neural network based interpolation for adaptive frame repair, dynamically choosing the interpolation depth based on the number of frames to be corrected. This enables reconstruction of the corrupted frames from the last valid and next valid keyframes. Our results show up to 1.4 point improvement in CLIP Score and up to 6.1 point improvement in warp error compared to SOTA baselines on the DAVIS and Pexels video datasets.
\end{abstract}


\section{Introduction}
\label{sec:intro}

Recent advances in artificial intelligence have significantly enhanced the quality, realism, and efficiency of synthetic image and video generation models~\cite{ho2022video, geyer2023tokenflow, munir2024three, liu2024sora}. These improvements have broadened applications in content creation, real-time video editing, and interactive media~\cite{kodaira2023streamdiffusion, croitoru2022diffusion, streamv2v}. Despite these strides, a critical yet often overlooked challenge persists, namely \textit{maintaining consistent object representation} across different video frames. This is important since subtle inconsistencies, including semantic drift, visual flickering, or transient artifacts, frequently arise during video synthesis and editing, diminishing the visual coherence and perceptual realism~\cite{geyer2023tokenflow}.

Current diffusion-based editing methods \cite{geyer2023tokenflow, mahmud2025ada, pnpDiffusion2023, meng2022sdedit}, predominantly use extended attention mechanisms to propagate information across frames to maintain temporal coherence. However, extending attention across multiple frames significantly increases the computational cost and memory requirements, often becoming prohibitively expensive\cite{geyer2023tokenflow, mahmud2025ada}. Moreover, these approaches do not provide formal guarantees for consistency, leaving room for errors that degrade the video quality.

To overcome these limitations, there is an emerging need for robust verification methods capable of \textit{provably} ensuring consistency between frames. Unlike methods relying solely on perceptual metrics which may still miss subtle inconsistencies or offer no formal assurances, a provable guarantee, such as that provided by a Satisfiability Modulo Theories (SMT) solver \cite{SMT_Intro}, can offer a mathematically-grounded assertion that specified consistency constraints (e.g., bounds on semantic feature drift) are met. This is crucial for detecting errors that evade heuristic checks.

\begin{figure*}[t] \centering \includegraphics[width=\linewidth]{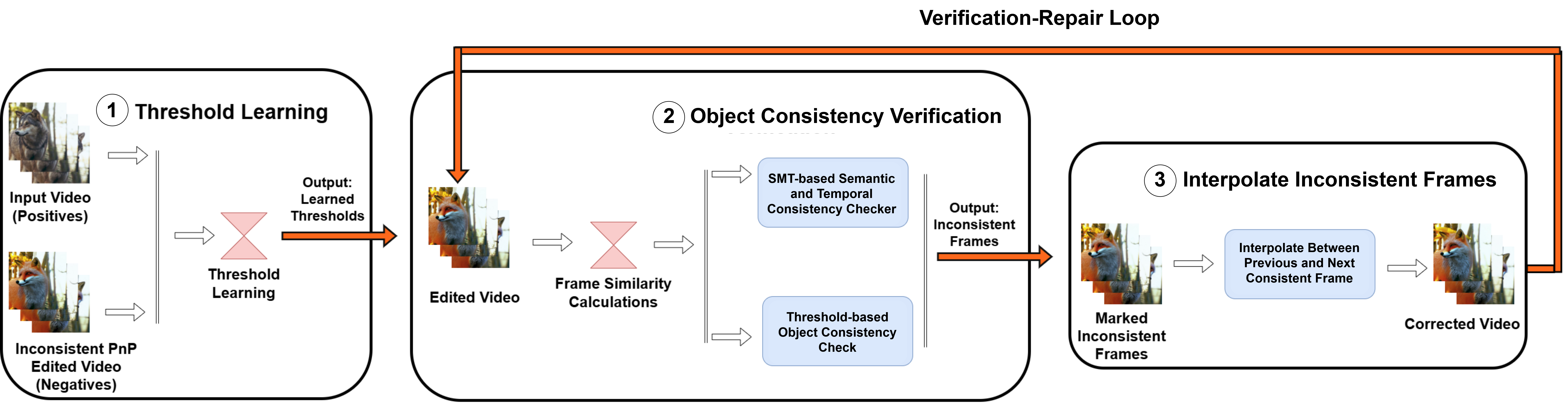} \caption{
\footnotesize{
\textbf{Overview of ObjectAlign.} \raisebox{.5pt}{\textcircled{\raisebox{-.9pt} {1}}} We first learn per‐metric consistency thresholds from “positive” original video clips and “negative” inconsistently edited clips.  
\raisebox{.5pt}{\textcircled{\raisebox{-.9pt} {2}}} Next, for each consecutive frame pair in a newly edited video, we compute semantic and perceptual similarities and apply both the learned threshold checks and an SMT‐based object consistency check on the embeddings to flag inconsistent transitions.  
\raisebox{.5pt}{\textcircled{\raisebox{-.9pt} {3}}} Finally, each contiguous block of flagged frames is repaired by adaptively interpolating between the nearest preceding and succeeding consistent keyframes, with the interpolation depth chosen according to the segment length.  The corrected frames can then be re‐verified in a closed loop until no inconsistencies remain.}
} \label{fig:overview}  \end{figure*}

In this paper, we propose \textbf{ObjectAlign}, a neuro-symbolic framework that can rigorously verify and adaptively repair object-level and temporal inconsistencies in edited video sequences. Our approach bridges perceptual metrics with symbolic verification techniques, ensuring both practical performance and formal consistency guarantees. To this end, we introduce three key contributions:

\begin{itemize}

\item First, we propose a new methodology that integrates multiple perceptual and semantic metrics, including CLIP-based semantic similarity, LPIPS perceptual distance, color histogram correlation, and segmentation-mask IoU, into a unified, \emph{learnable threshold-based classifier} for identifying object-level inconsistencies. By learning thresholds directly from data, our approach offers both flexibility and interpretability in inconsistency detection.

\item Second, we introduce a formal verification method to \textit{provably verify semantic and temporal consistency}. Specifically, we embed object features as constraints within a symbolic reasoning framework, enforcing per-dimension semantic bounds on masked CLIP embeddings. We also ensure temporal fidelity verification through a probabilistic model checker to verify the video satisfies a given temporal logic specification ($\Phi$). This ensures a mathematically grounded guarantee of semantic and temporal consistency within defined thresholds.

\item Finally, we develop an \textit{adaptive interpolation strategy} for correcting flagged inconsistencies. Our repair mechanism dynamically adjusts the interpolation depth based on the number of contiguous inconsistent frames identified, reconstructing corrupted frames from adjacent consistent keyframes, thus preserving a smooth temporal coherence.
\end{itemize}

\noindent Indeed, as shown in Figure~\ref{fig:overview}, ObjectAlign effectively integrates learnable perceptual metrics, formal semantic verification, and adaptive interpolation-based correction into a unified end-to-end pipeline. Our evaluation demonstrates that ObjectAlign reduces perceptual flickering and semantic drift, decreasing the warp error \cite{Warp_Error} from 107.4 to 101.3 compared to Plug and Play Diffusion (PnP) \cite{pnpDiffusion2023} on clips from the DAVIS \cite{davis} and Pexels \cite{pexels} video datasets.

The remainder of this paper is structured as follows: Section~\ref{sec:related_work} discusses related work in video synthesis and formal verification techniques. Section~\ref{sec:preliminaries} provides necessary background on diffusion models, perceptual metrics, and SMT solvers. Section~\ref{sec:method} describes the ObjectAlign methodology and technical innovations in detail. Experimental results and ablations are presented in Section~\ref{sec:results}. Finally, Section~\ref{sec:conclusion} summarizes our main contributions.


\section{Related work}
\label{sec:related_work}

\subsection{Video Editing and Object Consistency}

Recent works have explored training-free frameworks for improving or stylizing text-to-video generation by leveraging pre-trained text-to-image (T2I) models to edit video frames~\cite{Jeong2023GroundAVideoZG, khachatryan2023text2videozerotexttoimagediffusionmodels, Yang2023RerenderAV, wang2024zeroshotvideoeditingusing}. Approaches such as SDEdit~\cite{meng2022sdedit}, InstructPix2Pix~\cite{Brooks2022InstructPix2PixLT}, and ControlNet~\cite{zhang2023adding} provide general-purpose image editing capabilities that have been adapted for video by applying them frame-by-frame or with additional guidance. Several methods enhance video generation through refined text prompts~\cite{kim2024free2guidegradientfreepathintegral, luo2025enhanceavideobettergeneratedvideo}, or by combining text and image modalities for editing~\cite{zhang2024moonshot}. Plug-and-Play Diffusion~\cite{pnpDiffusion2023} and Free2Guide~\cite{kim2024free2guidegradientfreepathintegral} further enable flexible, training-free editing. Dreamix~\cite{molad2023dreamixvideodiffusionmodels} and Tune-A-Video~\cite{wu2022tuneavideo} demonstrate the use of video diffusion models and spatio-temporal tuning for improved consistency and style transfer. Real-time editing approaches such as StreamDiffusion and StreamV2V~\cite{kodaira2023streamdiffusion, streamv2v} enable efficient video editing.

A key limitation of these approaches lies in their difficulty to maintain temporal coherence and object consistency across frames. Methods like TokenFlow~\cite{geyer2023tokenflow}, Rerender~\cite{Yang2023RerenderAV}, and VideoP2P~\cite{Liu2023VideoP2PVE} address this by identifying key frames~\cite{Ceylan2023Pix2VideoVE} and propagating features across frames. Other approaches, such as Ground-A-Video~\cite{Jeong2023GroundAVideoZG}, FateZero~\cite{Qi2023FateZeroFA}, and Ada-VE~\cite{mahmud2025ada}, rectify cross-frame attention or integrate motion cues to improve consistency. Despite these advances, cross-frame attention remains computationally expensive and does not provide formal guarantees of consistency~\cite{geyer2023tokenflow}.

\subsection{Neuro-Symbolic Verification}

Neuro-symbolic methods aim at integrating the advancement of neural networks with the rigor of symbolic reasoning \cite{10.1093/nsr/nwac035, colelough2025neurosymbolicai2024systematic}. Neuro-symbolic methods use symbolic reasoning to provide formal guarantees in various domains. Specifically, in image and video synthesis, formal verification approaches such as SMT \cite{SMT_Intro} and temporal logic \cite{Choi_2024_ECCV,sp_2025_CVPR} can rigorously validate the consistency and semantic correctness of the generated content.

Recently, neuro-symbolic verification has been be explored for video searching, editing, and evaluation tasks \cite{sp_2025_CVPR, Choi_2024_ECCV}. Video classification employs graph-based relational modeling \citep{short1, short2}, while event detection leverages spatiotemporal pattern recognition in video streams \citep{event-detection-video-stream, cnn-event-detect, lstm-event-detect}. Neuro-symbolic frameworks enhance video question-answering \citep{neural-symbolic, chen2022comphy}, with applications extending to robotic action planning \citep{shoukry2017linear,hasanbeig2019reinforcement,kress2009temporal} and safety verification in autonomous driving systems \citep{jha2018safe,mehdipour2023formal}. These methods either construct graph structures \citep{graphical-model-relationship-detection, visual-symbolic, long3}, use latent-space representations as symbolic representations \citep{symbolic-high-speed-video, BertasiusWT21, neural-symbolic-cv}, or use formal language methods \citep{Baier2008} to design specifications.

In contrast with this prior work, our ObjectAlign uniquely combines learnable perceptual metrics with symbolic constraints using the SMT solving, thus providing formal guarantees for object consistency in video editing. Additionally, we complement our verification with adaptive interpolation to repair inconsistencies dynamically. ObjectAlign is the first work to explore object consistency correction by post-processing inconsistent frames identified by learnable perceptual metrics or formal verification.

\section{Preliminaries} \label{sec:preliminaries}

\subsection{Latent Diffusion Models}

Diffusion models~\citep{ho2020denoising, dhariwal2021diffusion} are generative models comprised of two main stochastic phases: (a) a \textit{forward process} that progressively adds noise to data, and (b) a \textit{reverse process} that learns to remove this noise to generate data.

The \textbf{forward process} is typically formulated as a fixed Markov chain that gradually introduces Gaussian noise to an initial data sample $\mathbf{z}_0$ over $T$ discrete time steps. If $\mathbf{z}_0$ is a sample from the true data distribution $p_\mathrm{data}$ (e.g., a clean image), this process yields a sequence of increasingly noisy samples $\mathbf{z}_1, \dots, \mathbf{z}_T$. The final sample $\mathbf{z}_T$ is ideally distributed according to a normal distribution, $\mathcal{N}(0, \mathbf{I})$, where $\mathbf{I}$ is the identity covariance matrix. The transition at each step $t$ is defined as:

\begin{equation}
q(\mathbf{z}_t \mid \mathbf{z}_{t-1}) = \mathcal{N}\bigl(\mathbf{z}_t;\,\sqrt{\alpha_t}\,\mathbf{z}_{t-1},\,(1-\alpha_t)\,\mathbf{I}\bigr),
\label{eq:forward_process}
\end{equation}

where $\mathbf{z}_{t-1}$ is the data sample at the previous time step, $\mathbf{z}_t$ is the sample at the current time step, and $\alpha_t$ is a parameter derived from a predefined noise schedule (e.g., $\alpha_t = 1-\beta_t\in (0, 1)$, where $\beta_t\in (0, 1)$ are small positive constants representing variance schedules).

The \textbf{reverse process} aims to reverse this noising procedure. It starts with a sample $\mathbf{z}_T \sim \mathcal{N}(0, \mathbf{I})$ and iteratively denoises it to produce a sample $\mathbf{z}_0$ that resembles data from the true distribution $p_\mathrm{data}$. This process is also a Markov chain, parameterized by a neural network with parameters $\theta$. The model is trained to predict the conditional probability distribution $p_\theta(\mathbf{z}_{t-1} \mid \mathbf{z}_t)$ for each step $t$:
\begin{equation}
p_\theta(\mathbf{z}_{t-1}\mid \mathbf{z}_t) = \mathcal{N}\bigl(\mathbf{z}_{t-1};\,\boldsymbol{\mu}_\theta(\mathbf{z}_t, t),\,\sigma_t^2\,\mathbf{I}\bigr),
\label{eq:reverse_process}
\end{equation}

where $\boldsymbol{\mu}_\theta(\mathbf{z}_t, t)$ is the mean of the Gaussian distribution for $\mathbf{z}_{t-1}$, predicted by the neural network conditioned on the noisy sample $\mathbf{z}_t$ and the time step $t$. The term $\sigma_t^2$ represents the variance at time step $t$, which is often predefined or learned as part of the noise schedule. The neural network is trained to make $p_\theta(\mathbf{z}_{t-1} \mid \mathbf{z}_t)$ accurately approximate the true posterior $q(\mathbf{z}_{t-1} \mid \mathbf{z}_t, \mathbf{z}_0)$.

\subsection{Metrics for Object Consistency Verification}
\label{subsec:consistency_metrics_formal}

To assess object consistency between video frames, ObjectAlign employs a combination of perceptual metrics and formal verification techniques. Let $f_i, f_j \in \mathcal{I}$ be two video frames, where $\mathcal{I}$ denotes the entire space of all video frames.

\paragraph{Perceptual Consistency Metrics.}
We utilize four established metrics to capture different aspects of visual and semantic similarity:

\begin{itemize}
    \item \textbf{Learned Perceptual Image Patch Similarity (LPIPS):} This metric quantifies low-level perceptual similarity. The LPIPS distance between any two frames $f_i$, $f_j$ is given by:
\begin{equation}
      \mathrm{LPIPS}(f_i,f_j) = \bigl\lVert \phi(f_i) - \phi(f_j)\bigr\rVert_2,
\end{equation}
    where $\phi\colon\mathcal{I}\to\mathbb{R}^d$ is a deep feature extractor~\cite{LPIPs} and $\|x\|_2$ denotes the Euclidean norm. Smaller LPIPS values indicate that the frames are more similar at a patch-perceptual level.
    \item \textbf{CLIP-based Semantic Similarity:} To measure high-level semantic alignment, we use the cosine similarity between image embeddings from a Contrastive Language-Image Pre-training (CLIP) model~\cite{CLIP_Score}:
\begin{equation}
      \mathrm{Sim}_{\mathrm{CLIP}}(f_i,f_j) = \frac{\langle e(f_i),\,e(f_j)\rangle}{\|e(f_i)\|\cdot\|e(f_j)\|_2},
\end{equation}
    where $e\colon\mathcal{I}\to\mathbb{R}^k$ is the CLIP image encoder, $\langle x, y\rangle$ denote the standard dot‐product and $\|x\|_2$ denotes the Euclidean norm. Values closer to $1$ signify stronger semantic correspondence between the frames.

    \item \textbf{Histogram Correlation:} To check for significant color shifts between frames, we compute the correlation between their color histograms. Let $h(f) \in \mathbb{R}^c$ be the flattened and normalized color histogram vector for frame $f$, where $c$ represents the dimensionality. The histogram correlation is:
\begin{equation}
      \mathrm{Sim}_{\mathrm{Hist}}(f_i,f_j) = \frac{h(f_i)^\top h(f_j)}{\|h(f_i)\|_2 \cdot \|h(f_j)\|_2},
\end{equation}
    where $(\cdot)^\top$ denotes transpose. Values closer to $1$ indicate higher similarity in the overall color distributions.

    \item \textbf{Mask IoU:} For object-level geometric consistency, we compute the IoU of foreground object masks. Let $M\colon\mathcal{I}\to\{0,1\}^{H\times W}$ be the binary foreground mask obtained for a frame (e.g., via the Segment Anything Model (SAM)~\cite{SAM_Model}), where $H$ and $W$ are the frame height and width. The IoU is: 
\begin{equation}
      \mathrm{IoU}\bigl(M(f_i),M(f_j)\bigr) = \frac{\lvert M(f_i)\cap M(f_j)\rvert}{\lvert M(f_i)\cup M(f_j)\rvert}.
\end{equation}
    This value, ranging from $0$ to $1$, quantifies the spatial overlap of the primary objects.
\end{itemize}

These perceptual metrics provide complementary empirical checks on object consistency, covering low-level appearance, high-level semantic content, color distribution, and object geometry, respectively. However, taken alone, they cannot inherently provide formal guarantees of coherence.

\paragraph{Formal Verification with SMT Solvers.}
To address the limitations of purely metric-based approaches and introduce rigorous consistency checks, ObjectAlign provides formal verification using SMT solvers. SMT solvers can determine the satisfiability of logical formulas with respect to background theories, enabling us to enforce provable bounds on specific features. In our context, we use an SMT solver to enforce \emph{semantic stability} by asserting bounds on the object drift. We use object masks $M_i$ to compute separate embeddings for the foreground object $e(f_i, M_i)$ and the background $e(f_i, \neg M_i)$. We then use an SMT solver to formally verify a \textit{conjunctive} formula that ensures both \textbf{object identity stability} and \textbf{background stability}:
\begin{equation}
\small
\begin{split}
& \left( \forall j \, \bigl|\,e_j(f_i, M_i) - e_j(f_{i+1}, M_{i+1})\,\bigr| \le \epsilon_s \right) \land \\
& \left( \forall j \, \bigl|\,e_j(f_i, \neg M_i) - e_j(f_{i+1}, \neg M_{i+1})\,\bigr| \le \epsilon_{bg} \right)
\end{split}
\label{eq:smt_constraint_prelim}
\end{equation}
where $\epsilon_s$ and $\epsilon_{bg}$ are semantic drift tolerances. An SMT solver checks if this set of constraints is satisfiable; if it is, then we have a formal guarantee that no individual semantic feature dimension has drifted beyond the specified tolerances $\epsilon_s$ and $\epsilon_{bg}$. ObjectAlign leverages this neuro-symbolic verification to complement the aforementioned learned perceptual metrics, providing a more robust and reliable consistency assessment than using the perceptual metrics alone.


\section{Proposed Methodology}
\label{sec:method}

ObjectAlign consists of three stages executed in a closed
verification--repair loop (Fig.~\ref{fig:overview}):
\emph{\raisebox{.5pt}{\textcircled{\raisebox{-.9pt} {1}}}~metric–based scoring with learned thresholds},  
\emph{\raisebox{.5pt}{\textcircled{\raisebox{-.9pt} {2}}}~neuro-symbolic consistency verification}, and  
\emph{\raisebox{.5pt}{\textcircled{\raisebox{-.9pt} {3}}}~adaptive frame repair via neural interpolation}.
The loop repeats until every neighbouring frame pair
satisfies \emph{all} consistency constraints.

\subsection{Inconsistency identification (Step {\textcircled{\raisebox{-.9pt} {1}}} in Fig.~\ref{fig:overview})}
\label{subsec:inconsistency_identification}

\subsubsection{Metric Based Consistency Scoring}
\label{subsec:metric_scoring}

\paragraph{Feature vector.}
For two consecutive frames $f_i,f_{i+1}$, we extract  
(\emph{a}) cosine similarity of CLIP embeddings 
$S_{\cos}$,  
(\emph{b}) color–histogram correlation $S_{\text{hist}}$,  
(\emph{c}) mask–IoU $S_{\text{iou}}$, and  
(\emph{d}) perceptual distance $D_{\text{lpips}}$.  
We invert LPIPS so that \emph{larger} values denote higher
consistency, i.e.\ $\tilde S_{\text{lpips}} = -D_{\text{lpips}}$.
Hence the feature vector is
\(
\mathbf{s}_i = \bigl[\,S_{\cos},\,
                     S_{\text{hist}},\,
                     S_{\text{iou}},\,
                     \tilde S_{\text{lpips}}\bigr]^{\!\top}.
\)
These specific metrics are chosen for their complementary strengths in assessing frame-to-frame object consistency: CLIP \cite{CLIP_Score} similarity captures high-level semantic content alignment, LPIPS \cite{LPIPs} evaluates low-level perceptual appearance, color histogram correlation checks for drastic color shifts, and mask IoU quantifies object-level geometric overlap and spatial stability, thereby providing a comprehensive empirical check as noted in Section~\ref{sec:preliminaries}.

\paragraph{Learnable thresholds.}
We treat each dimension of the feature vector $\mathbf{s}_i$ independently and learn a threshold vector $\boldsymbol{\tau} = \bigl[\tau_{\cos}, \tau_{\text{hist}}, \tau_{\text{iou}}, \tau_{\text{lpips}}\bigr]^{\!\top}$ from a small \emph{positive} set $\mathcal{P}$ (adjacent frames from the unedited video) and a \emph{negative} set $\mathcal{N}$ (pairs of original vs. edited frames, considered inconsistent). For each frame pair $i$, we compute the element-wise difference vector $\Delta_i$ between its feature vector $\mathbf{s}_i$ and the learned threshold vector $\boldsymbol{\tau}$: $\Delta_i = \mathbf{s}_i - \boldsymbol{\tau}$. The probability that a pair is consistent for a single threshold $k$ ($P_k(i)$) is modeled by the sigmoid function:
\begin{equation}
    P_k(i)
      \;=\;
      \sigma\!\bigl(\lambda\,\Delta_k\bigr), 
      \quad 
      \sigma(z)=\frac{1}{1+e^{-z}},
\end{equation}
where $\lambda$ is a sharpness constant. The four thresholds in $\boldsymbol{\tau}$ are simultaneously optimized by minimizing the binary cross-entropy loss:
\begin{equation}
\begin{split}
    \mathcal{L}_{\text{BCE}} = & -\frac{1}{|\mathcal{P}|+|\mathcal{N}|} \sum_{i \in \mathcal{P}\cup\mathcal{N}} \Bigl[ y_i \log(\textit{P}_\textit{k}(i)) \\
    & \qquad + (1-y_i) \log(1-\textit{P}_\textit{k}(i)) \Bigr]
\end{split}
\end{equation}

where $y_i = 1$ for $i \in \mathcal{P}$ and $y_i = 0$ for $i \in \mathcal{N}$. Optimization is performed using Adam \cite{Adam}.

\subsubsection{Neuro-Symbolic Verification (Step {\textcircled{\raisebox{-.9pt} {2}}} in Fig.~\ref{fig:overview})}
\label{subsec:smt}

While the metric classifier is effective in practice for capturing perceptual inconsistencies, it offers no \emph{formal} guarantee against all forms of object drift, particularly subtle semantic shifts that may fall within learned perceptual thresholds but still represent a logical inconsistency. The SMT-based verification step (see \raisebox{.5pt}{\textcircled{\raisebox{-.9pt} {2}}} in Fig~\ref{fig:overview}) addresses this by combining low-level feature stability with high-level temporal fidelity.

Given that the scalar perceptual metrics (e.g., $S_{\text{hist}}$) are directly evaluated against their learned thresholds (Eq.~\eqref{eq:final_consistency}), SMT verification is reserved for the high-dimensional CLIP embeddings to enforce semantic stability. We therefore impose an SMT constraint on the \textbf{masked CLIP embeddings} (introduced in Sec.~\ref{subsec:consistency_metrics_formal}). Specifically, we verify the stability of both the foreground object $e(f, M)$ and the background $e(f, \neg M)$ independently, defining this semantic stability constraint as $\mathcal{C}_{\text{neuro}}$:
\begin{equation}
\label{eq:neuro_constraint}
\small
\begin{split}
\mathcal{C}_{\text{neuro}} \equiv & \left( \forall j \, \bigl|\,e_j(f_i, M_i) - e_j(f_{i+1}, M_{i+1})\,\bigr| \le \epsilon_s \right) \land \\
& \left( \forall j \, \bigl|\,e_j(f_i, \neg M_i) - e_j(f_{i+1}, \neg M_{i+1})\,\bigr| \le \epsilon_{bg} \right)
\end{split}
\end{equation}

We complement this stability check with a high-level temporal fidelity metric \citep{sp_2025_CVPR, choi2025wellfixpostimproving}. This component calculates a \textit{satisfaction probability} by verifying the video's formal representation (automaton $\mathcal{A_{\nu}}$) against the text prompt's temporal logic specification ($\Phi$) using a probabilistic model checker function, $\Psi$. A video is considered formally verified \textit{only if} it satisfies both the low-level stability constraints ($\mathcal{C}_{\text{neuro}}$) and the high-level temporal requirements. We define this unified neuro-symbolic constraint, $\mathcal{P}_{\text{formal}}$, as the logical conjunction of these two conditions:
\begin{equation}
\mathcal{P}_{\text{formal}} \equiv \mathcal{C}_{\text{neuro}} \land \left( \Psi(\mathcal{A_{\nu}},\Phi) \ge \tau \right)
\label{eq:formal_constraint}
\end{equation}
Here, $\mathcal{P}_{\text{formal}}$ is satisfied if and only if the SMT solver finds the frame-to-frame drift constraints $\mathcal{C}_{\text{neuro}}$ (the first conjunct) satisfiable for all frames, \textit{and} the probabilistic model checker finds that the temporal fidelity $\Psi(\mathcal{A_{\nu}},\Phi)$ meets or exceeds a specified probability threshold $\tau$ (the second conjunct).

\subsubsection{Joint Consistency Criterion} \label{subsubsec:Joint_Consistency}

A transition is declared \emph{consistent} ($C(i)=1$) if and only if (iff)
\emph{all} thresholds are satisfied simultaneously \emph{and} the formal constraints are met:

\begin{equation}
\label{eq:final_consistency}
\small 
\begin{split}
    C(i) = & \Bigl(S_{\cos} \ge \tau_{\cos}\Bigr)\land
             \Bigl(S_{\text{hist}} \ge \tau_{\text{hist}}\Bigr)\land \\
           & \Bigl(S_{\text{iou}} \ge \tau_{\text{iou}}\Bigr)\land
             \Bigl(D_{\text{lpips}} \le \tau_{\text{lpips}}\Bigr)\land \\
           & \Bigl(P_\text{formal}(i)=1\Bigr)
\end{split}
\end{equation}
All indices with $C(i)=0$ form the inconsistent set
$\mathcal{I}$.

\textbf{Formal Consistency Guarantees.} The joint consistency criterion $C(i)$ defined in Eq.~\eqref{eq:final_consistency} combines the learned threshold checks with a formal SMT constraint. Our joint consistency criterion assures that a frame‐pair declared consistent by our pipeline ($C(i)=1$) exhibits a bounded drift according to every metric and the formal semantic check included in our criteria. These formal bounds underpin ObjectAlign’s robustness in improving video consistency.

\begin{algorithm}
\caption{\textsc{ObjectAlign} verification–repair loop}
\label{alg:objectalign}
\small
\begin{algorithmic}[1]
\State \textbf{Input:} edited video $V=\{f_0,\dots,f_{T-1}\}$ \Comment{$T$ is the total number of frames in video $V$.}
\State Learn thresholds $\boldsymbol{\tau}$ on positive and negative set 
        \Comment{Sec.~\ref{subsec:inconsistency_identification}}
\Repeat
    \State $\mathcal{I}\gets\emptyset$
    \For{$i=0$ \textbf{to} $T-2$}  \Comment{Iterate over all $T-1$ frame transitions $(f_i, f_{i+1})$}
        \State compute $P_\text{metric}(i)$, $P_\text{formal}(i)$
        \If{$C(i)$ (Eq.~\ref{eq:final_consistency}) is \textbf{false}}
            \State $\mathcal{I}\gets\mathcal{I}\cup\{i\}$
        \EndIf
    \EndFor
    \For{each contiguous run $[i_s,i_e]\subseteq\mathcal{I}$} \Comment{$i_s, i_e$: start/end indices of a run of inconsistent transitions}
        \State $k\gets i_e-i_s+1$ \Comment{Number of frames $f_{i_s}, \dots, f_{i_e}$ to repair}
        \State replace $f_{i_s},\dots,f_{i_e}$ with
               corrected frames  \hfill
               \Comment{Sec.~\ref{subsec:repair}}
    \EndFor
\Until{$\mathcal{I}=\emptyset$}
\State \textbf{return} verified \& corrected video $V$
\end{algorithmic}
\end{algorithm}

\subsection{Adaptive Frame Repair (Step {\textcircled{\raisebox{-.9pt} {3}}} in Fig.~\ref{fig:overview})}
\label{subsec:repair}

The adaptive frame repair stage (see \raisebox{.5pt}{\textcircled{\raisebox{-.9pt} {3}}} in Fig~\ref{fig:overview}) fundamentally relies on the presence or eventual emergence of consistent anchor frames surrounding any block of identified inconsistencies. This principle can be conceptualized using Linear Temporal Logic (LTL)~\cite{Manna}. 

Let $AP_{IB}$ be an atomic proposition that is true when a contiguous block of frames is currently identified as an \emph{`InconsistentBlock`} requiring repair. Let $AP_{CAB}$ be true if a \emph{`ConsistentAnchorBefore`} (i.e., a suitable frame $f_{i_s-1}$) exists or is established, and $AP_{CAA}$ be true if a \emph{`ConsistentAnchorAfter`} (i.e., a suitable frame $f_{i_e+1}$) exists or is established. The iterative verification-repair loop of ObjectAlign (Algorithm~\ref{alg:objectalign}) operates under the premise that the video sequence will eventually satisfy the property:
\begin{equation}
\label{eq:ltl_repair_assumption}
\always (AP_{IB} \imply (\eventually AP_{CAB} \andlogic \eventually AP_{CAA}))\
\end{equation}

This LTL formula asserts that it is always ($\always$) the case that if an inconsistent block requiring repair ($AP_{IB}$) exists, then eventually ($\eventually$) a consistent anchor frame will be found or established before it ($AP_{CAB}$), and eventually ($\eventually$) a consistent anchor frame will be found or established after it ($AP_{CAA}$), thus enabling interpolation. Our framework aims to progressively achieve this state, allowing for repair even when initial edits contain extended inconsistent segments.

\begin{figure*}[th]
  \centering
  \includegraphics[width=\textwidth]{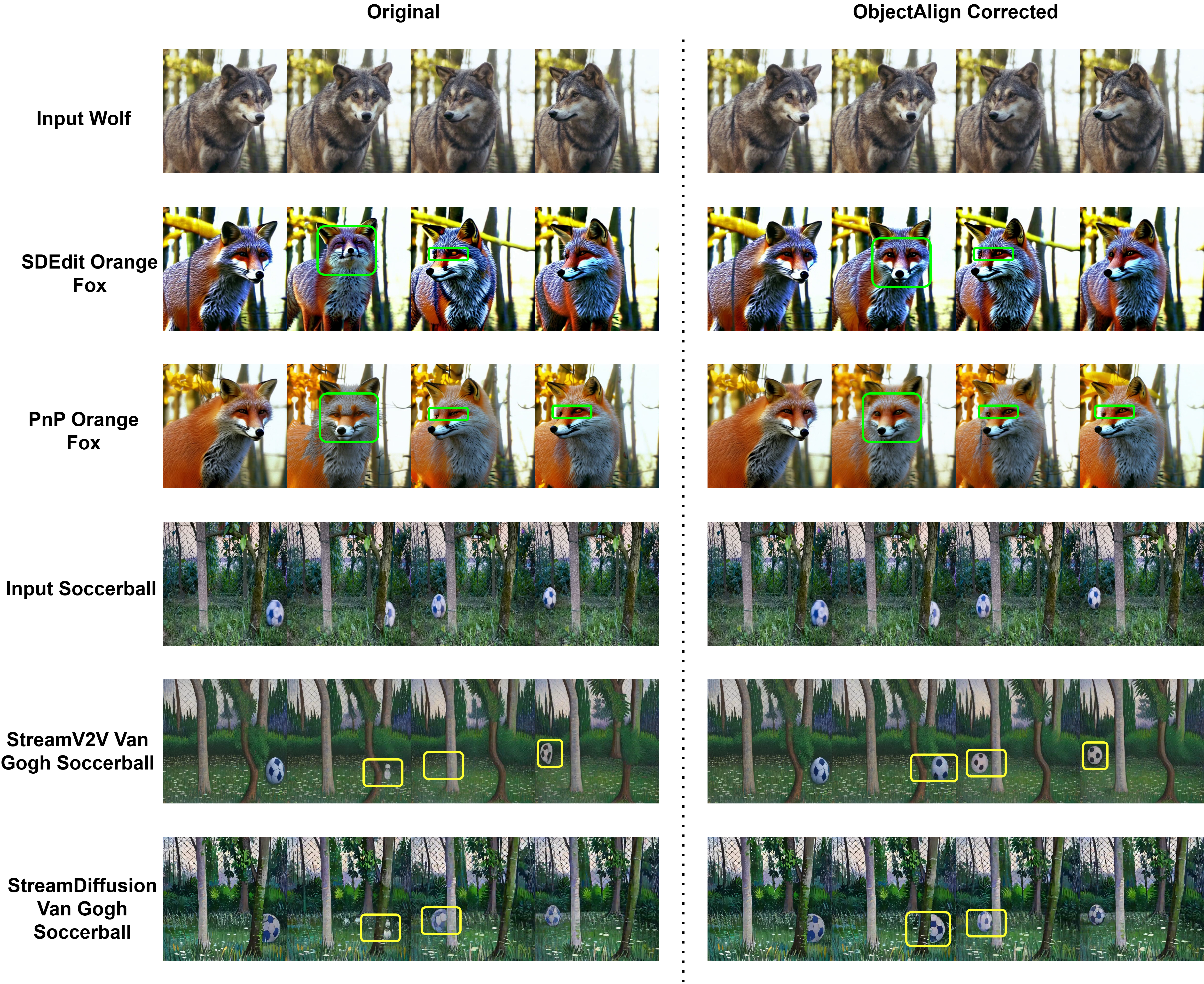}
     \caption{\footnotesize{\textbf{Qualitative comparison of ObjectAlign corrections across different editing pipelines.} Before ObjectAlign correction (left), both SDEdit and PnP in the "Orange Fox" edits incorrectly alter the wolf's shape and color across consecutive frames (\textit{highlighted in green boxes}). Similarly, StreamV2V and StreamDiffusion in the "Van Gogh Soccerball" edits cause the soccerball to intermittently disappear and reappear (\textit{highlighted in yellow boxes}). These inconsistencies are accompanied by noticeable color and style drift, perceptual flickering, and identity misalignment. After applying ObjectAlign (right), these issues are effectively mitigated, resulting in greater semantic and temporal consistency.}}
  \label{fig:qual_results}
\end{figure*}

Given a contiguous sequence of frames marked as inconsistent, we apply adaptive neural network based interpolation using RIFE~\cite{huang2022rife}. Let $\bigl[i_s,i_e\bigr]\!\subseteq\!\mathcal{I}$ represent a maximal run of $k=i_e - i_s + 1$ inconsistent frames. To reconstruct these frames, we first identify the closest consistent frames immediately preceding and following this sequence: $f_{i_s-1}$ and $f_{i_e+1}$ (whose existence is anticipated by the property in Eq.~\eqref{eq:ltl_repair_assumption}). We then dynamically select the interpolation depth ($\gamma$) as a function of the number of frames needing repair ($k$), defined by: $ \gamma=\lceil\log_2(k+1)\rceil $. This adaptive interpolation depth ensures that longer runs of inconsistencies are addressed with deeper interpolation, generating sufficient intermediate frames to preserve smooth and coherent motion. The repaired frames replace the original inconsistent frames, producing an updated, more consistent video sequence.

The ObjectAlign pipeline re-runs Sections~\ref{subsec:inconsistency_identification} to~\ref{subsec:repair} on $V$ until $\mathcal{I}=\emptyset$\, as shown in Algorithm \ref{alg:objectalign}. ObjectAlign is able to provide both \emph{empirical} quality and \emph{formal} consistency guarantees through its neuro-symbolic verification pipeline.


\section{Experimental Results}
\label{sec:results}

\subsection{Experimental Setup}\label{subsec:setup}

Our experiments utilize a dataset of 95 text-to-video prompt pairs obtained from the DAVIS~\cite{davis} and Pexels~\cite{pexels} datasets. The prompts include both manually designed descriptions and those inspired from \cite{pnpDiffusion2023, streamv2v}. The videos cover a diverse array of subjects including animals and humans. The prompts encompass various scenarios involving object edits, style changes, and structural modifications. Our primary baselines for comparison are PnP \cite{pnpDiffusion2023}, SDEdit \cite{meng2022sdedit}, StreamV2V \cite{streamv2v}, and StreamDiffusion \cite{kodaira2023streamdiffusion} without our object consistency corrections. Evaluations focus on object-level consistency and semantic fidelity using established metrics such as CLIP \cite{CLIP_Score} scores and Warp Error~\cite{Warp_Error}. All experiments were performed on one Nvidia 3090 GPU.

\begin{table*}[th]
    \centering
    \caption{\footnotesize{\textbf{ObjectAlign Improvements.} Comparison of video quality and consistency metrics \textit{before} (Original) and \textit{after} applying ObjectAlign (Edited) to videos processed by different base editing methods (PnP, SDEdit, StreamV2V, and StreamDiffusion). Scores are shown for different edit categories, with \textcolor{teal}{(+Improvement)} indicating the improvement attributed to ObjectAlign. Higher scores are better for CLIP Score and VBench \cite{VBench} motion smoothness, subject consistency, and background consistency metrics; lower is better for Warp Error.}}
    \label{tab:objectalign_results} 
    \renewcommand{\arraystretch}{1.3}
    \setlength{\tabcolsep}{3pt} 
    \resizebox{\linewidth}{!}{
        \begin{tabular}{cccccccccc} 
            \toprule
            \multicolumn{2}{c}{\multirow{2}{*}{\textbf{Edit Category}}} & \multicolumn{2}{c}{\textbf{PnP~\cite{pnpDiffusion2023}}} & \multicolumn{2}{c}{\textbf{SDEdit~\cite{meng2022sdedit}}} & \multicolumn{2}{c}{\textbf{StreamV2V~\cite{streamv2v}}} & \multicolumn{2}{c}{\textbf{StreamDiffusion~\cite{kodaira2023streamdiffusion}}} \\
            \cmidrule(lr){3-4} \cmidrule(lr){5-6} \cmidrule(lr){7-8} \cmidrule(lr){9-10}
            &  & Orig & Edited & Orig & Edited & Orig & Edited & Orig & Edited \\
            \midrule
            \multirow{5}{*}{\textbf{Object Edits}}
                & CLIP Score $\uparrow$ & 97.0 & 98.1 \textcolor{teal}{(+1.1)} & 96.7 & 98.1 \textcolor{teal}{(+1.4)} & 97.1 & 97.9 \textcolor{teal}{(+0.8)} & 95.2 & 96.0 \textcolor{teal}{(+0.8)} \\
                & Warp Error $\downarrow$ & 107.4 & 101.3 \textcolor{teal}{(+6.1)} & 105.5 & 100.8 \textcolor{teal}{(+4.7)} & 100.5 & 98.8 \textcolor{teal}{(+1.7)} & 108.7 & 103.5 \textcolor{teal}{(+5.2)} \\
                & Motion Smoothness $\uparrow$ & 0.917 & 0.930 \textcolor{teal}{(+0.013)} & 0.903 & 0.925 \textcolor{teal}{(+0.022)} & 0.916 & 0.935 \textcolor{teal}{(+0.019)} & 0.887 & 0.901 \textcolor{teal}{(+0.014)} \\
                & Subject Consistency $\uparrow$ & 0.913 & 0.925 \textcolor{teal}{(+0.012)} & 0.900 & 0.915 \textcolor{teal}{(+0.015)} & 0.920 & 0.931 \textcolor{teal}{(+0.011)} & 0.884 & 0.899 \textcolor{teal}{(+0.015)} \\
                & Background Consistency $\uparrow$ & 0.921 & 0.925 \textcolor{teal}{(+0.004)} & 0.904 & 0.917 \textcolor{teal}{(+0.013)} & 0.917 & 0.924 \textcolor{teal}{(+0.007)} & 0.892 & 0.908 \textcolor{teal}{(+0.016)} \\
            \midrule
            \multirow{5}{*}{\textbf{Style Edits}} 
                & CLIP Score $\uparrow$ & 97.6 & 98.2 \textcolor{teal}{(+0.6)} & 97.3 & 98.0 \textcolor{teal}{(+0.7)} & 97.5 & 97.9 \textcolor{teal}{(+0.4)} & 95.8 & 96.4 \textcolor{teal}{(+0.6)} \\
                & Warp Error $\downarrow$ & 106.3 & 101.6 \textcolor{teal}{(+4.7)} & 105.3 & 100.5 \textcolor{teal}{(+4.8)} & 99.5 & 98.6 \textcolor{teal}{(+0.9)} & 107.2 & 103.3 \textcolor{teal}{(+3.9)} \\
                & Motion Smoothness $\uparrow$ & 0.938 & 0.973 \textcolor{teal}{(+0.035)} & 0.930 & 0.958 \textcolor{teal}{(+0.028)} & 0.933 & 0.940 \textcolor{teal}{(+0.007)} & 0.920 & 0.936 \textcolor{teal}{(+0.016)} \\
                & Subject Consistency $\uparrow$ & 0.905 & 0.937 \textcolor{teal}{(+0.032)} & 0.904 & 0.925 \textcolor{teal}{(+0.021)} & 0.912 & 0.928 \textcolor{teal}{(+0.016)} & 0.896 & 0.908 \textcolor{teal}{(+0.012)} \\
                & Background Consistency $\uparrow$ & 0.913 & 0.932 \textcolor{teal}{(+0.019)} & 0.903 & 0.920 \textcolor{teal}{(+0.017)} & 0.915 & 0.928 \textcolor{teal}{(+0.013)} & 0.900 & 0.913 \textcolor{teal}{(+0.013)} \\
            \midrule
            \multirow{5}{*}{\textbf{Overall Average}} 
                & CLIP Score $\uparrow$ & 97.3 & 98.2 \textcolor{teal}{(+0.9)} & 97.0 & 98.1 \textcolor{teal}{(+1.1)} & 97.3 & 97.9 \textcolor{teal}{(+0.6)} & 95.5 & 96.2 \textcolor{teal}{(+0.7)} \\
                & Warp Error $\downarrow$ & 106.9 & 101.5 \textcolor{teal}{(+5.4)} & 105.4 & 100.7 \textcolor{teal}{(+4.7)} & 100.0 & 98.7 \textcolor{teal}{(+1.3)} & 108.0 & 103.4 \textcolor{teal}{(+4.6)} \\
                & Motion Smoothness $\uparrow$ & 0.928 & 0.952 \textcolor{teal}{(+0.024)} & 0.917 & 0.942 \textcolor{teal}{(+0.025)} & 0.925 & 0.938 \textcolor{teal}{(+0.013)} & 0.904 & 0.919 \textcolor{teal}{(+0.015)} \\
                & Subject Consistency $\uparrow$ & 0.909 & 0.931 \textcolor{teal}{(+0.022)} & 0.902 & 0.920 \textcolor{teal}{(+0.018)} & 0.916 & 0.930 \textcolor{teal}{(+0.014)} & 0.890 & 0.904 \textcolor{teal}{(+0.014)} \\
                & Background Consistency $\uparrow$ & 0.917 & 0.929 \textcolor{teal}{(+0.012)} & 0.905 & 0.919 \textcolor{teal}{(+0.014)} & 0.916 & 0.926 \textcolor{teal}{(+0.010)} & 0.896 & 0.911 \textcolor{teal}{(+0.015)} \\
            \bottomrule
        \end{tabular}
    }
\end{table*}

\subsection{Qualitative Results}\label{subsec:qual_results}

Figure~\ref{fig:qual_results} presents visual comparisons between ObjectAlign and the baseline PnP \cite{pnpDiffusion2023} and SDEdit \cite{meng2022sdedit} methods on multiple challenging scenarios. We observe that ObjectAlign significantly reduces perceptual flicker, artifact generation, and object drift. In particular, ObjectAlign effectively maintains stable object identities across frames, producing results noticeably smoother and more temporally coherent than baseline methods.

\subsection{Quantitative Results}\label{subsec:quan_results}

\paragraph{CLIP Score.}
We measure semantic consistency using the CLIP similarity score~\cite{CLIP_Score}, defined as the cosine similarity of CLIP embeddings between consecutive frames. Higher scores reflect greater semantic stability ($\uparrow$). When ObjectAlign is applied to correct the outputs of various base editing methods, it consistently enhances semantic stability. For instance, drawing from the "Overall Average" results in Table~\ref{tab:objectalign_results}, applying ObjectAlign to videos edited by PnP improves the CLIP Score from an original \(97.3\) to \(98.2\). For SDEdit, the score increases from \(97.0\) to \(98.1\); for StreamV2V~\cite{streamv2v}, it improves from \(97.3\) to \(97.9\); and for StreamDiffusion~\cite{kodaira2023streamdiffusion}, the score is enhanced from an original \(95.5\) to \(96.2\). This demonstrates that ObjectAlign effectively preserves or improves semantic content preservation across frames when applied to a range of editing techniques.

\paragraph{Warp Error.}
Temporal coherence is evaluated via Warp Error~\cite{Warp_Error}, which computes pixel-wise discrepancies after warping edited frames by the original video’s optical flow. Lower Warp Error indicates greater temporal consistency ($\downarrow$). When applied to various base editing methods, ObjectAlign consistently reduces their Warp Error, thereby enhancing temporal coherence. For instance, as detailed in Table~\ref{tab:objectalign_results} (Overall Average section), ObjectAlign improves the Warp Error for PnP from an original score of \(106.9\) down to \(101.5\). Similarly, for SDEdit, the error is reduced from \(105.4\) to \(100.7\); for StreamV2V, from \(100.0\) to \(98.7\); and for StreamDiffusion, from \(108.0\) to \(103.4\). These results confirm that our method produces more temporally consistent videos when used to correct the outputs of established editing techniques.

\paragraph{VBench Perceptual Metrics.}
To further assess video quality across diverse perceptual dimensions, we employ metrics from the VBench benchmark~\cite{VBench}, specifically Motion Smoothness, Subject Consistency, and Background Consistency. For these metrics, higher scores are preferable ($\uparrow$). As shown in Table~\ref{tab:objectalign_results}, ObjectAlign consistently improves these scores when applied to the outputs of different editing methods across both object and style edit categories. For instance, when ObjectAlign is applied to videos edited using PnP, the Motion Smoothness score increases from \(0.928\) to \(\mathbf{0.952}\), and Subject Consistency improves from \(0.909\) to \(\mathbf{0.931}\). Similarly, for a baseline like StreamDiffusion, ObjectAlign enhances Motion Smoothness from \(0.904\) to \(\mathbf{0.919}\) and Subject Consistency from \(0.890\) to \(\mathbf{0.904}\). These examples, representative of the broader findings in Table~\ref{tab:objectalign_results}, indicate enhanced visual quality in terms of smoother motion, more stable subject appearance, and more coherent backgrounds. Further details on the VBench benchmark can be found in the original VBench documentation~\cite{VBench}.

\subsection{Ablation Studies}\label{subsec:ablations}

\paragraph{Ablation Study on Diffusion-Based Inpainting for Frame Repair.}
To evaluate the efficacy of our adaptive interpolation for frame repair (Section~\ref{subsec:repair}), we conducted an ablation study comparing it against an alternative approach using a pre-trained Stable Diffusion (SD) inpainting pipeline~\cite{Stable_Diffusion}. For this experiment, inconsistent frame outputs of PnP \cite{pnpDiffusion2023} were targeted for repair. Segmentation masks obtained via SAM~\cite{SAM_Model} from a consistent reference frame guided the inpainting region, and textual prompts were provided. As shown in Table~\ref{tab:ablation_repair_methods}, the SD inpainting method yielded minimal beneficial impact on key metrics such as CLIP Score and Warp Error when applied to repair inconsistent PnP outputs, improving them by only 0.1 points. In contrast, ObjectAlign's interpolation demonstrates substantial improvements on CLIP Score \cite{CLIP_Score} and warp error \cite{Warp_Error} for the same PnP outputs. These findings support our choice of targeted interpolation from consistent anchor frames for frame repair.

\begin{table}[ht]
    \centering
    \footnotesize
    \caption{\footnotesize{\textbf{Ablation Study: Frame Repair Methods for PnP Outputs.} Comparison of ObjectAlign's RIFE-based interpolation against Stable Diffusion (SD) inpainting for repairing frames from the PnP baseline. "Original" refers to PnP output before repair. SD Inpainting refers to scores after the inpainting based repair method. Improvements \textcolor{teal}{(+Value)} are relative to PnP (Original). The ablation study is performed over 18 edited video sequences.}}
    \label{tab:ablation_repair_methods}
    \renewcommand{\arraystretch}{1.2}
    \setlength{\tabcolsep}{4pt} 
        \begin{tabular}{@{}lccc@{}}
            \toprule
            Metric & Original & SD Inpainting & Interpolation (ObjectAlign)\\
            \cmidrule(lr){2-2} \cmidrule(lr){3-3} \cmidrule(lr){4-4}
            \midrule
            CLIP Score $\uparrow$ & 97.0 & 97.1 \textcolor{teal}{(+0.1)} & 97.7 \textcolor{teal}{(+0.7)} \\
            Warp Error $\downarrow$ & 106.4 & 106.6 \textcolor{red}{(-0.2)} & 100.3 \textcolor{teal}{(+6.1)} \\
            \bottomrule
        \end{tabular}
\end{table}

\paragraph{Ablation Study on Verification Checks.}

We ablate the impact of individual consistency verification checks—semantic similarity (CLIP cosine), perceptual similarity (LPIPS), histogram correlation, object-mask overlap (IoU), and the SMT-based semantic embedding constraint—on identifying inconsistent frames. This study is performed over 35 edited video sequences of lengths between 40 and 280 frames each. As detailed in Table~\ref{tab:ablation_checks}, we observe that the IoU-based object-mask consistency check is most frequently triggered (22.3\% of total frames), reflecting its sensitivity to spatial discrepancies in segmentation masks. The SMT-based embedding constraint triggers second most often (16.6\%), underscoring the benefit of formal semantic bounds. The perceptual LPIPS check triggers third (15.5\%), highlighting its effectiveness at detecting subtle visual artifacts. Histogram correlation and CLIP-based semantic similarity check flag inconsistencies less frequently (8.7\% and 7.1\%, respectively), indicating that color and global semantic shifts are comparatively rarer. Overall, the combined use of complementary verification checks ensures robust detection of diverse inconsistency types, each targeting different aspects of perceptual, spatial, and semantic coherence.

\begin{table}[h]
\centering
\footnotesize
\def\arraystretch{1.2}
\setlength{\tabcolsep}{3pt}
\caption{\footnotesize{\textbf{Ablation on individual consistency verification checks.} We report the percentage of frames flagged as inconsistent by each verification check over all sequences. Higher percentage indicates greater sensitivity of the check in detecting inconsistencies.}}
\label{tab:ablation_checks}
\begin{tabular}{@{}lccccc@{}}
\toprule
\textbf{Verification Check}      & IoU   & SMT   & LPIPS & Histogram & CLIP Cosine \\ 
\midrule
Percentage flagged (\%) & 22.3 & 16.6  & 15.5  & 8.7       & 7.1         \\ 
\bottomrule
\end{tabular}
\end{table}


\section{Conclusion}
\label{sec:conclusion}

In this paper, we have introduced ObjectAlign, a neuro-symbolic framework designed to detect, formally verify, and adaptively correct object-level inconsistencies in edited video sequences. Our approach integrates learnable perceptual metrics, neuro-symbolic verification, and adaptive neural network based interpolation to ensure semantic fidelity, temporal fidelity, and visual coherence. 

Experimental evaluations demonstrate ObjectAlign's ability to substantially reduce semantic drift and visual artifacts, achieving superior performance in both perceptual consistency and temporal coherence, compared to existing baseline methods. Furthermore, ablation studies confirm the importance of each component in our design, highlighting the effectiveness of combining learnable consistency thresholds, symbolic reasoning, and adaptive interpolation. ObjectAlign thus represents an important step towards provably consistent and visually stable video editing.

\clearpage

{
    \small
    \bibliographystyle{ieeenat_fullname}
    \bibliography{main}
}

\input{sec/X_suppl}

\end{document}

%% file: sec/X_suppl.tex
\clearpage
\maketitlesupplementary

\theoremstyle{plain} 
\newtheorem{theorem_app}{Proposition}[section] 
\newtheorem{assumption_app}{Assumption}[section]
\newtheorem{lemma_app}{Lemma}[section]
\newtheorem{proposition_app}{Proposition}[section]
\newtheorem{corollary_app}{Corollary}[section]

\theoremstyle{definition} 
\newtheorem{definition_app}{Definition}[section]
\newtheorem{example_app}{Example}[section]

\theoremstyle{remark} 
\newtheorem{remark_app}{Remark}[section]

\section{User Study}
\label{app:user_study}

\begin{figure}[htp]
\centering
\includegraphics[width=0.95\columnwidth]{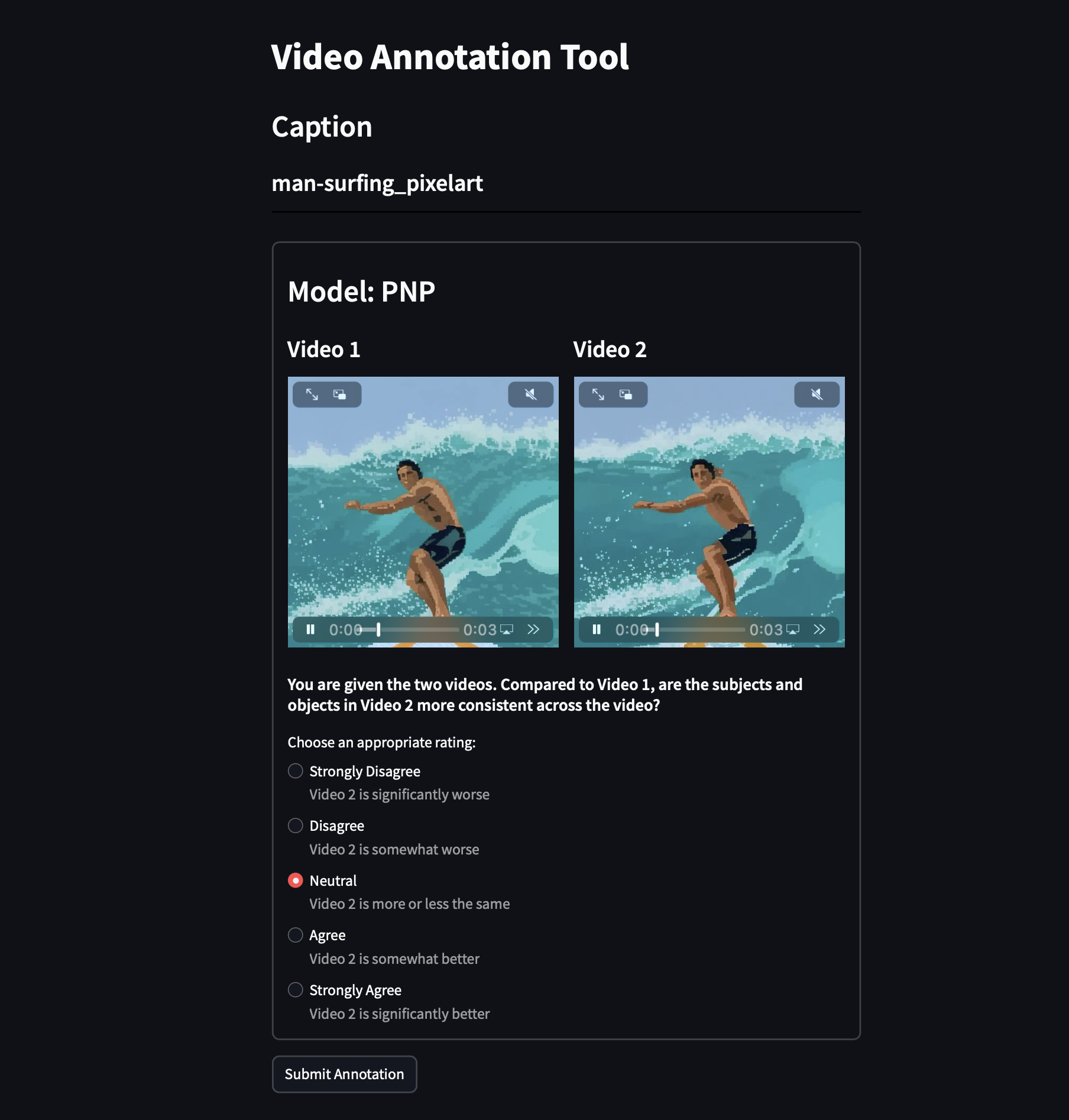}
\caption{\textbf{Annotation tool for User Study.} Participants are asked to evaluate the efficacy of ObjectAlign in terms of correcting videos to improve subject consistency. We provide a randomized base video and an edited video whose presentation order is randomized to remove bias, and users are asked to compare whether Video 2 is better than Video 1.}
\label{fig:annotation_tool}
\end{figure}

We conducted a user study with 10 participants who assessed 32 video pairs generated from four different baseline editing methods: PnP \cite{pnpDiffusion2023}, SDEdit \cite{meng2022sdedit}, StreamDiffusion\cite{kodaira2023streamdiffusion}, and StreamV2V \cite{streamv2v}. Of note, our user study includes more participants than StreamV2V (3 participants) \cite{streamv2v}, FlowVid (5 participants) \cite{liang2023flowvid}, and the same number of participants as ADA-VE (10 participants) \cite{mahmud2025ada}. Each pair comprised an original edited video from a baseline model and the corresponding video corrected by our proposed \textbf{ObjectAlign} method. The presentation order of the videos was randomized. Participants were asked to rate their level of agreement with the statement: ``\emph{video 2 is better in terms of the consistency of the subjects across the video compared to video 1.}'' The responses were scored from 1 (\emph{Strongly Disagree}) to 5 (\emph{Strongly Agree}). We show the user interface in Figure \ref{fig:annotation_tool}.

\begin{figure}[h!]
\centering
\includegraphics[width=\columnwidth]{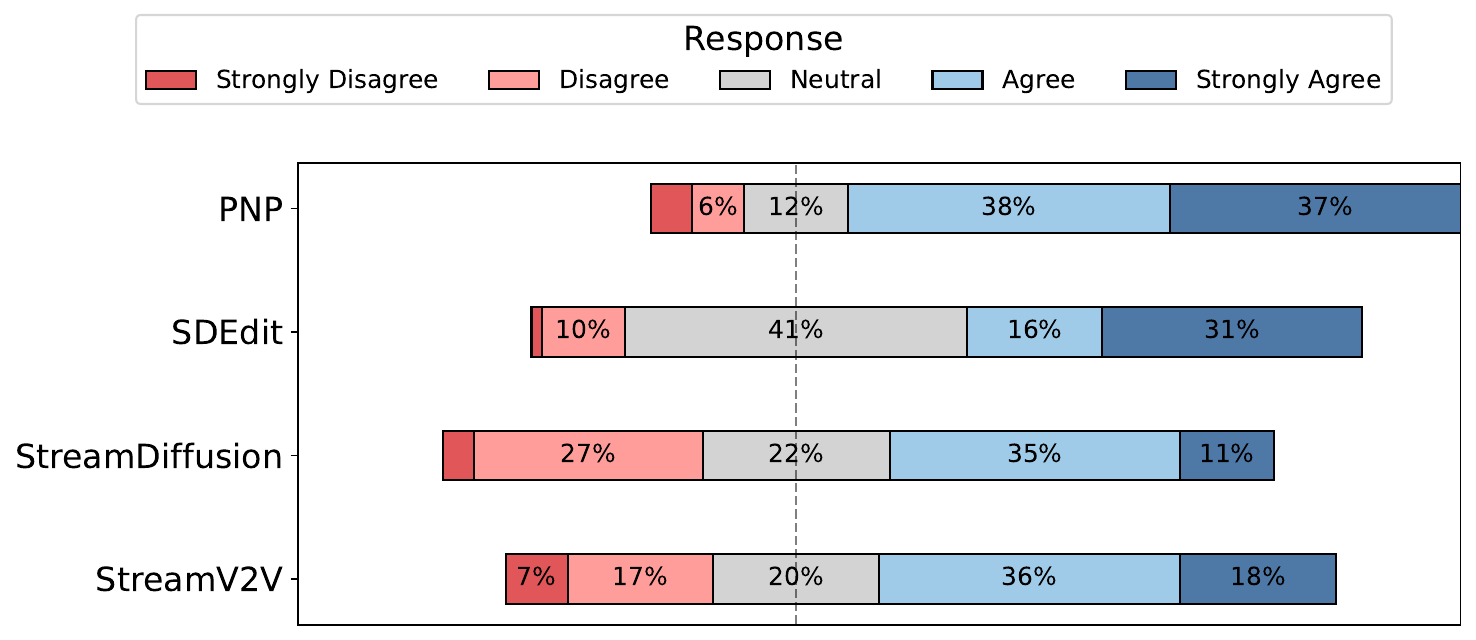}
\caption{\textbf{User Study Results on Perceptual Improvement by ObjectAlign.} Participants were asked to evaluate whether the ObjectAlign corrected videos demonstrate noticeable improvements in subject consistency compared to baseline edited videos (users did not know which video is the original and which was the ObjectAlign corrected version). Responses ranged from ``\emph{Strongly Disagree}'' to ``\emph{Strongly Agree}''. Results indicate a clear user preference for ObjectAlign corrected videos, especially prominent in the PnP method, where 75\% of participants expressed strong agreement or agreement. Conversely, StreamDiffusion corrections showed the lowest perceived improvement, indicating variations in ObjectAlign's effectiveness depending on the underlying editing pipeline, but improvements compared to the baseline regardless.}
\label{fig:user_study_results}
\end{figure}

Figure~\ref{fig:user_study_results} summarizes the distribution of participant responses aggregated per baseline method. The results clearly demonstrate a strong user preference for videos corrected by ObjectAlign across all editing baselines. In particular, for the PnP method, 75\% of participants either agreed or strongly agreed that ObjectAlign significantly improved subject consistency. Overall, the user study confirms that ObjectAlign consistently enhances perceptual quality by effectively addressing artifacts and inconsistencies introduced by existing video editing methods.

These qualitative user insights complement the quantitative evaluations and further validate ObjectAlign's practical benefits for improving temporal coherence and object consistency in edited videos.

\clearpage

\section{Additional Qualitative Results}
\label{app:qual_results}

\begin{figure*}[th!]
\centering
\includegraphics[width=\textwidth]{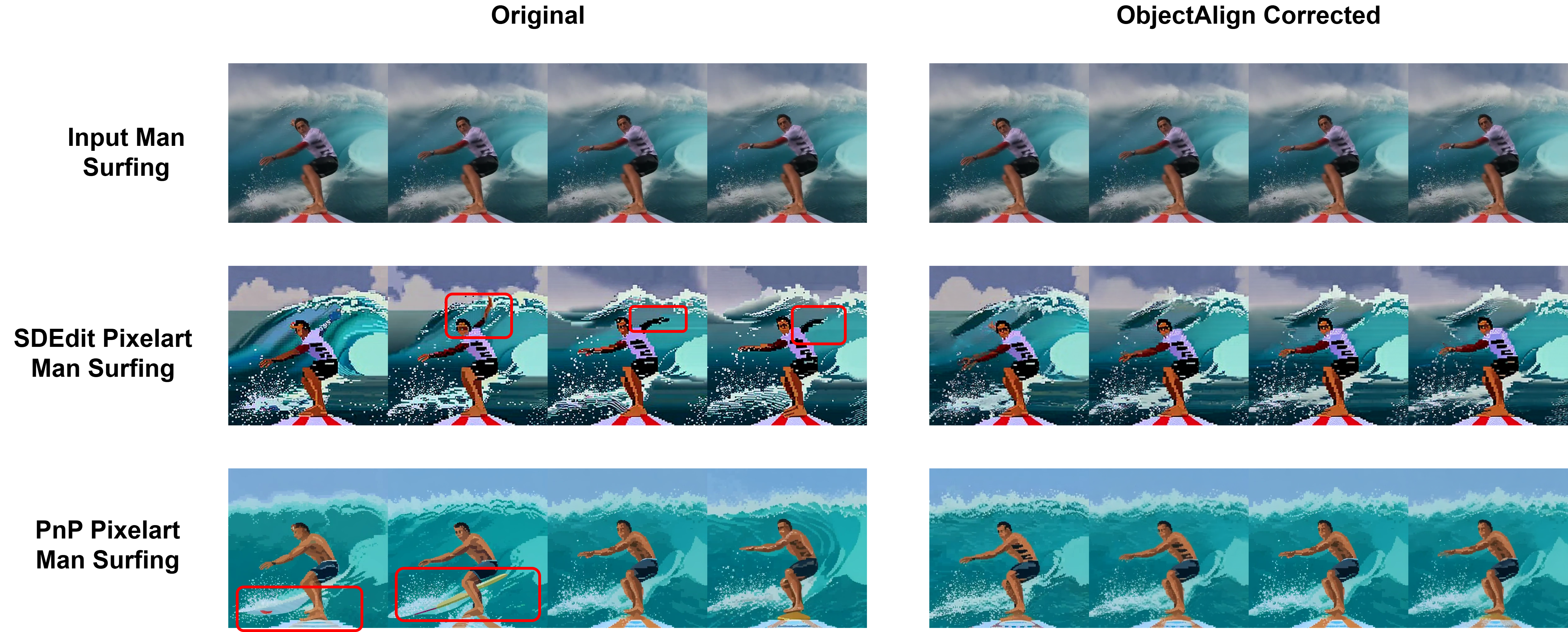}
\caption{\textbf{Further Qualitative Comparisons of ObjectAlign Corrections.}
(Top Row) Original real-world input video depicting a man surfing.
(Middle Row, SDEdit Pixelart) Pixelart stylized frames produced by SDEdit introduce transient artifacts and distortions (highlighted in red boxes) around the surfer's arm. ObjectAlign correction successfully removes these artifacts, ensuring temporal consistency of object shapes.
(Bottom Row, PnP Pixelart) PnP Pixelart stylization introduces significant spatial inconsistencies in the surfer's surfboard and introduces mysterious artifacts as highlighted in the red boxes. The far left frame shows a red artifact, and the second frame from the left introduces a random yellow object passing through the surfer. ObjectAlign effectively corrects these inconsistencies, resulting in a smoother and visually coherent video.}
\label{fig:suppl_qual_results}
\end{figure*}

Beyond the examples presented in the main paper (Figure~\ref{fig:qual_results}), this section provides further qualitative evidence of ObjectAlign's effectiveness in correcting inconsistencies introduced by various video editing pipelines. 

Figure~\ref{fig:suppl_qual_results} showcases additional challenging scenarios, specifically comparing pixelart stylizations produced by SDEdit \cite{meng2022sdedit} and PnP \cite{pnpDiffusion2023} against their ObjectAlign corrected counterparts. Red bounding boxes highlight notable artifacts and temporal inconsistencies in the edited sequences, such as distortions in the surfer's arm (SDEdit, middle row) and structural inconsistencies around the surfer's board and random objects appearing (PnP, bottom row). 

The ObjectAlign corrected sequences (right columns) effectively resolve these issues, demonstrating improved stability in object identity and shape. These visual results complement our quantitative evaluations, further confirming that ObjectAlign robustly enhances temporal coherence and semantic consistency in diverse video editing scenarios.

\section{Runtime Efficiency}
\label{app:runtime}
The runtime overhead introduced by ObjectAlign is minimal, requiring approximately $3$\% additional computation time compared to baseline PnP \cite{pnpDiffusion2023} editing, and $4$\% additional runtime compared to SDEdit \cite{meng2022sdedit}. The runtime overhead is primarily due to adaptive interpolation and SMT-based verification. The average processing time per frame remains acceptable for practical use, enabling ObjectAlign integration into existing image and video-editing workflows. Overall, ObjectAlign achieves a superior balance between semantic consistency, temporal coherence, and computational efficiency compared to existing methods.

\section{Limitations and Future Work}
\label{app:limitations}

While ObjectAlign effectively identifies and corrects object-level inconsistencies introduced during video editing, it remains inherently constrained by the quality of the underlying frame edits. Specifically, ObjectAlign relies on the existence of sufficient ``consistent'' keyframes to interpolate between them and reconstruct corrupted frames. If the edited frames are uniformly poor, such as containing pervasive visual artifacts or severe semantic drift throughout the video, then ObjectAlign cannot effectively recover a consistent video sequence, as no valid anchor points exist for interpolation. We leave this issue for future work.

Despite this limitation, ObjectAlign consistently improves temporal coherence and semantic consistency when applied atop capable editing pipelines, demonstrating clear benefits in real-world editing scenarios. 